\documentclass[journal]{IEEEtran}

\ifCLASSINFOpdf
   \usepackage[pdftex]{graphicx}
   \graphicspath{{../pdf/}{../jpg/}}
   \DeclareGraphicsExtensions{.pdf,.jpeg,.png}
\else
   \usepackage[dvips]{graphicx}
   \graphicspath{{../eps/}}
   \DeclareGraphicsExtensions{.eps}
\fi
\usepackage{multirow}

\usepackage{amsmath}
\usepackage{algorithmic}
\ifCLASSOPTIONcompsoc
  \usepackage[caption=false,font=normalsize,labelfont=sf,textfont=sf]{subfig}
\else
  \usepackage[caption=false,font=footnotesize]{subfig}
\fi
\begin{document}
%
\title{Multi-appearance Segmentation and Extended 0-1 Program for Dense Small Object Tracking}
%
%
%

\author{Longtao~Chen,~\
		Jing~Lou,~\
		Wei~Zhu,~\
		Qingyuan~Xia,~\
        Mingwu~Ren~\
\thanks{L. Chen, J. Lou, W. Zhu and M. Ren are with the School of Computer Science and Engineering, Nanjing University of Science and Technology(NUST), Nanjing XiaoLingWei 200, China. Corresponding author is M. Ren, E-mail: {mingwuren@163.com}}
}

%
%

\markboth{Journal of \LaTeX\ Class Files,~Vol.~14, No.~8, July~2016}%
{Shell \MakeLowercase{\textit{et al.}}: Bare Demo of IEEEtran.cls for IEEE Journals}
%



\maketitle

\begin{abstract}

Aiming to address the fast multi-object tracking for dense small object in the cluster background, we review track orientated multi-hypothesis tracking(TOMHT) with consideration of batch optimization. Employing autocorrelation based motion score test and staged hypotheses merging approach, we build our homologous hypothesis generation and management method. A new one-to-many constraint is proposed and applied to tackle the track exclusions during complex occlusions. Besides, to achieve better results, we develop a multi-appearance segmentation for detection, which exploits tree-like topological information and realizes one threshold for one object. Experimental results verify the strength of our methods, indicating speed and performance advantages of our tracker.
\end{abstract}

\begin{IEEEkeywords}
multi-object tracking, small object tracking, small object detection, track-oriented multi-hypothesis tracking.
\end{IEEEkeywords}

%
\IEEEpeerreviewmaketitle

\section{Introduction}
%
%
%
%
\IEEEPARstart{I}{n} video content analysis, whether for interpretation, indexing or coding, tracks of objects are of much importance. One of the important fields is dim small object tracking. Detection and tracking of small and dim moving objects are increasingly becoming vital for scenes like military infrared guidance, physical particles analysis and micro-animal observation. The objective of this paper is to develop algorithms than can detect and track small object in the complex scenario. This algorithm should be capable of accurately locating dim small object, starting and maintaining path and terminating it.

Many challenging factors stand in the way of successful tracking processing. It may occur events (temporary misdetection, occlusions, crossings), from which important ambiguities in the association of consecutive measurements to a track can arise. Proper localization for small object is the first challenge. Variable intensity of object with time, structured backgrounds, electronic noise and frequent occlusions are some examples of factors impeding the detection of small object. The overall unreliability of object detection results in corrupted measurements. The other factor comes from the data association. Due to the featureless character, limited by objects' similar small and isotropic shapes, the spatial position usually is the only feature to be relied on for data association. Thus, object motion should be extensively exploited to provide valued information.

\subsection{Discussion of association category}
Generally, the core problem of multi-object tracking is believed to be the data association. Data association determines the source that each measurement derives from, in other words, to build the very link between the measurement and track. From this aspect, methods of most contemporary multi-object tracking could be divided into two main categories.

One-to-one association, also known as unique neighbour association, means that each measurement should attach to up to only one track. One measurement could belong to an existing track, otherwise is regarded as the start of a new track or a false alarm. In short, it results from a unique source. The classic methods inspired by this rule include plain NN (nearest neighbour) and its modified variant, for instance GNN(global nearest neighbour). Those early methods immediately make the association decision after acquiring new measurements. In fact, delayed decision will improve the reliability of association for the sake of integrating later information. This idea just postpones the association decision for several frames, utilizing the information from subsequent measurements, and had been proved to considerably effective. MHT \cite{Blackman1999Design,Demos1990Applications,BarshalomMHTTAES,MHTBlackmanAESM}, MDA \cite{MDATAES}, SGTS \cite{PolynomialassociationTAEC,SGTSTAEC} are the typical methods of such idea. Reid proposed the original MHT algorithm \cite{ReidMHTTAC}. Afterwards, many researchers followed the work and explored the power of this algorithm by employing efficient assignment methods,such as A* search \cite{MHTBlackmanAESM}, Murty's \cite{COXMHTPAMI}, and optimizing the framework. MDA \cite{MDATAES}, stand for multiple dimensional assignment, expands the bipartite graph matching in MHT to multiply dimensional assignment. SGTS employs a semi-greedy algorithm to get the approximately optimal solution of association assumption.

Those improved one-to-one association methods with delayed decision also bring many new issues. The most serious one is association hypothesis explosion, which comes from the exponentially growth of combination of association. For now, this exponential disaster is partially alleviated by all kinds of restriction and pruning technologies for branches.

Moreover, most network optimization based methods could be classified into this category. They often utilize an abstract connected graph to represent the tracking problem. Since they would always add an one-to-one association constraint on the optimization problem, they satisfy the definition of one-to-one association method. Those methods include spatio-temporal path based optimization like K-Shortest paths optimization\cite{kshortestpami} proposed by J\'{e}r\^{o}me Berclaz \textit{et al.}, and tracklet based optimization like the works of Bing Wang \textit{et al.} \cite{TrackletPAMI,WangMetricICCV}. Actually, network optimization based tracking have been a crucial research focus especially in the last fewer decades. Many works' impressive results \cite{networkflowCVPR,GlobaloptiGreedyCVPR2011,LagrangianRelaxationCVPR2013} have indicated that it's a useful method with simple and clear framework. Zhang \textit{et al.} \cite{networkflowCVPR} used a push-relabel method to solve the min-cost flow problem. J\'{e}r\^{o}me Berclaz \textit{et al.} and Pirsiavash  \textit{et al.} \cite{GlobaloptiGreedyCVPR2011} proposed to use more efficient successive shortest path algorithms, which can provide roughly the same globally optimal tracking results with less running time. Butt \textit{et al.} \cite{LagrangianRelaxationCVPR2013} incorporated higher-order track smoothness constraints for multi-target tracking.

The second category is many-to-one association(also called all-neighbours association), where multiple measurements are used for the update of certain one track. The principal assumption is the case that each measurement within the threshold gate should contribute to the update of track, but with different weights. This method naturally avoids the fixed association as well as exponential combination. The typical methods include PDA and JPDA \cite{JPDA1}. The PDA algorithm calculates the association probabilities to the target being tracked for each validated measurement at the current time. This probabilistic or Bayesian information is used in the PDAF tracking algorithm, to account for the measurement origin uncertainty. JPDA could only track objects with fixed and informed number.

Beyond one-to-one (measurement-to-track) association and many-to-one association, we exploit one-to-many association to solve approximative motion occlusions. Some studies \cite{ExperimentMHT,evaluationstressfulTAES} show that the precision of JPDA may be inferior to MHT in some cases with massive objects. After thoughtful analysis of MHT, to maintain its ability in dealing with the massive low observable objects meanwhile avoid the daunting computations and time consumption, a new tracking framework is developed. The MMHT(management for multi-hypothesis tree) inspired by the TOMHT is employed as the generator of plausible hypothesis. Then, extended 0-1 program is used for hypothesis selection, which integrates considerations of the mutual exclusion and other critical facts.

\subsection{motivation}
The complexity is a fatal defect of MHT as well as related variations. MHT had been proved to be an effective method compared with other methods in dense object occasion. However, massive objects and exponentially growing scale of hypothesis form the nearly intractable problem to get compatible hypothesis solutions during probability evolution. Generally, this compatible sets problem would be transferred as a graph problem. Then, clustering divides it into some individual problems, and each individual is formed as a linear program or MWISP problem, etc. Besides, those computation about compatible hypothesis sets will be performed at each frame and so come with the frequent massive computation. Sometimes, those clustering processing and compatible sets searching include similar graph structure over fewer frames, which brings partly repeated computation. Because most of the incompatible relations, caused by crossings or something else, are inherited from the last frame to record history events. No each new frame brings new incompatible events, hence the compatible graph remains quite similar with last one sometimes.

To address massive computational complexity of MHT especially in dense scene, we try to extend the interval between graph processes. Besides, familiar graph structure is formed within a small interval, our effort could help to partly eliminate such phenomenon. Generally, trajectories can be generated online\cite{onlineconquer2015ICCV,Sanchez2016Online}, offline\cite{WangMetricICCV,MilanDetectionTrajectoryExclusionCVPR2013,ChariPairwiseNetworkflowCVPR2015}, or with a short latency\cite{PoiesiDensityTCSVT2015,ChoiNearonlineALF_ICCV2015}. Batch optimization tracking is in some sense like deferred tracking with affordable delay, as long as short batch length is applied. We try to employ batch decision to MHT framework.

The first problem we encountered is exponential explosion of hypothesis. The result of graph processing determines whether a hypothesis would be pruned or maintained. Prolonging interval means adding the depth of hypothesis tree. A new and strong hypothesis management method is needed to restrict the number of hypotheses and reserve the valid ones in the meantime. Then, the second problem is hypothesis selection. With more hypotheses, graph processing based algorithm may not be suited to handle it since the graph may expand to a larger degree. Increasing graph scale exponentially expands solution space.

Quite a number of segmentation algorithms have been used for the detection of small object, and proved to be highly efficient. Local contrast method proposed by C.L.Philip Chen \textit{et al.} \cite{localcontrastdetTGRS} showcases state-of-art abilities. However, its defect, to expand the object for the sake of using maxpool operation, is not good for accurate segmentation. In fact, most detectors concentrate on enhancing emergence probability of small objects. They usually don't think over for cases of dense object with lots of occlusions. To achieve better tracking results, we propose our multi-appearance segmentation. Unlike normal segmentation often utilizing unified threshold for single image, multi-appearance segmentation adopts different thresholds for different objects in the same image. To distinguish touching objects during occlusions, we need to use different combinations of thresholds. A topological tree structure is built to organize the relationship between objects under different thresholds.

We make the following contributions: (1)We propose a one-to-many association based constraint for dense small object tracking, and implement this constraint by extended 0-1 program, to the best of our knowledge this new association idea is the first that differ to common practices; (2)To maintain the hypothesis set in a tractable scale, we design a MMHT method for hypothesis management, which makes deeper tree steerable. (3)A novel multi-appearance segmentation method is proposed for small object detection. It utilizes topological tree structures to management the relationship between local thresholds for different objects, and refines individual thresholds. (4)Owning to the efforts to reduce the complexity and number of hypotheses, the implementation of our tracker is proved to be impressively fast.  

\begin{figure*}[!t]
	\centering
	\includegraphics[width=7.16in]{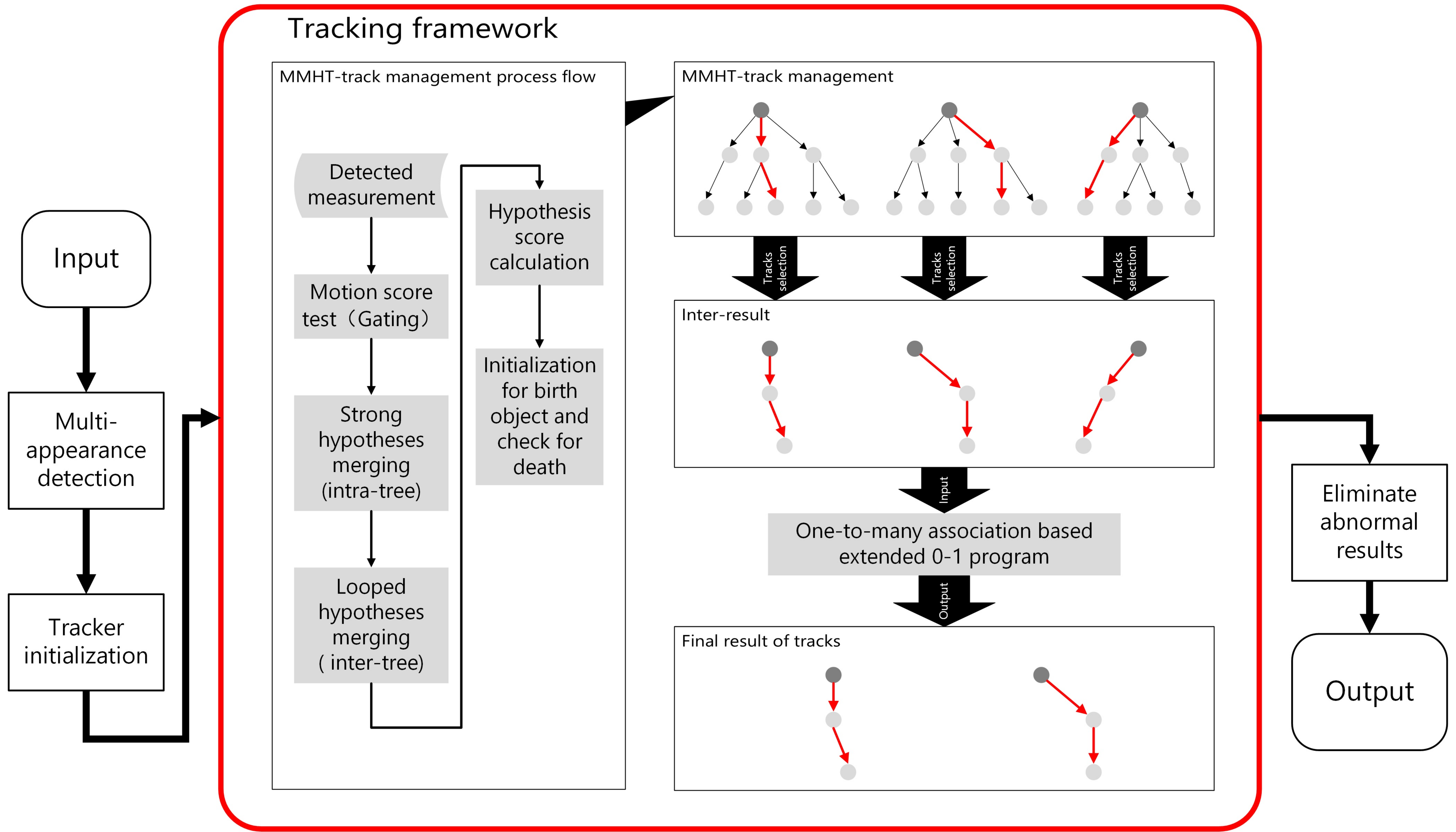}
	\caption{Our proposed framework.}
	\label{fig_whole_frame}
\end{figure*}\textbf{}

\subsection{Outline of the Paper}
The primary methods proposed in this paper will include two parts, presented in section 2 and 3. In section 2, we propose the multi-appearance segmentation for small object detection. Then, our tracking method will be presented in section 3. In section 4, the experiments about detection, tracking and verifying new constraint will be introduced. Finally, we discuss our results and future work in section 5.  


\begin{figure}[!t]
	\centering
	\includegraphics[width=1.8in]{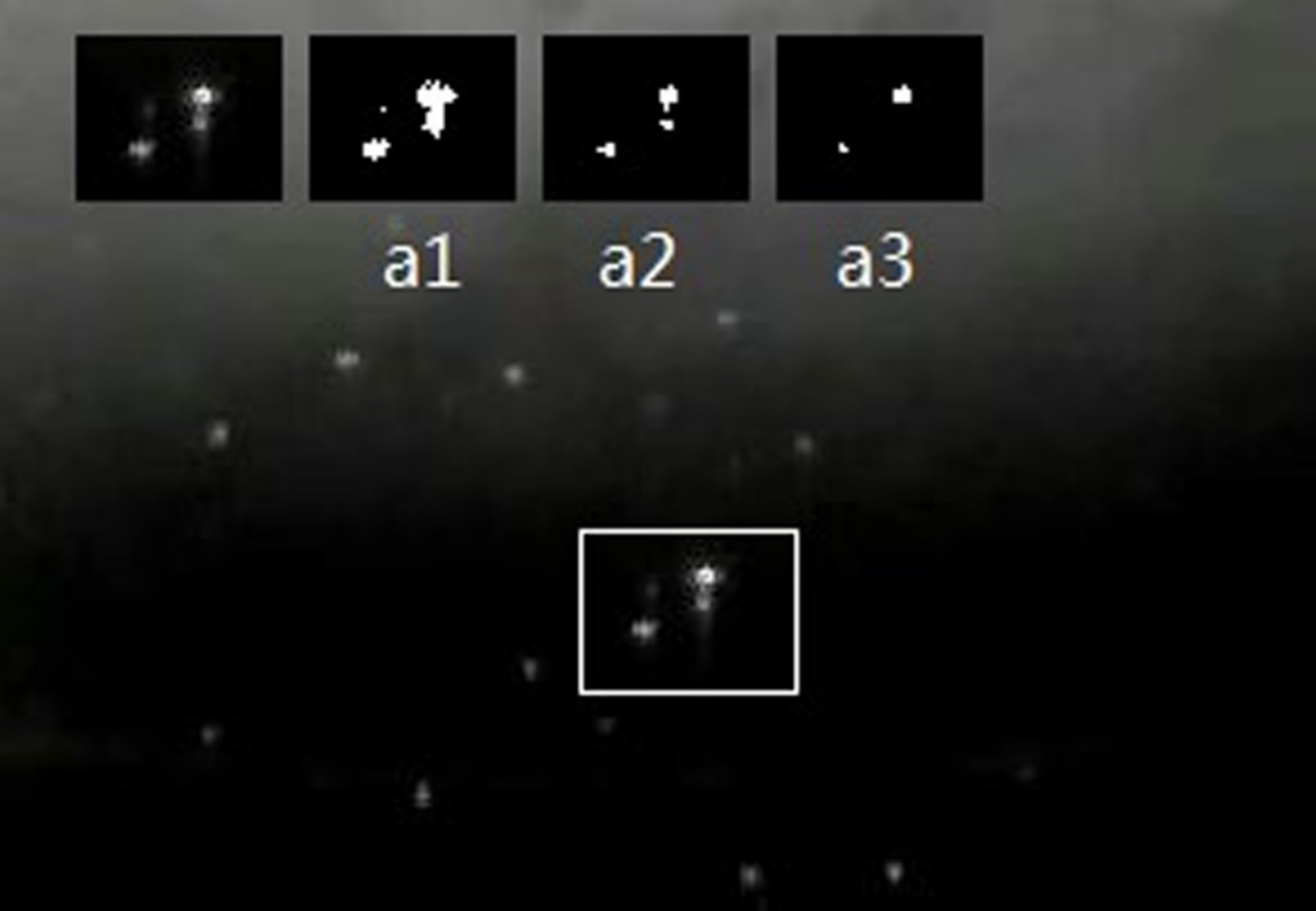}
	\caption{Example of image slides. Threshold in $a1$ makes the right objects connected to each other, which merges them to a single object. As for threshold of $a3$, segmentation with too high threshold directly loses sight of right-down object. Besides, the segmentation of $a2$ loses sight of left-up object.}
	\label{fig_multi_slide}
\end{figure}
\section{Multi-appearance segmentaion based detection method}
\subsection{Multi-appearance segmentaion}
We employ the same paradigm as tracking-by-detection framework, which is popularly used in visual tracking \cite{kshortestpami,JiangLinearprgramCVPR2007,WUEfficienttrackCVPR2011,GlobaloptiGreedyCVPR2011,networkflowCVPR,LiHybirdBoostedCVPR2009}. So detection of small object is our first task, and it is obviously a fundamental and vital step. Small objects usually appear as points or irregular blocks of few pixels, which certainly could not contain much information except intensity and rough shape. In an image with low SNR, objects are totally mixed up with noises, so plausibly that it's always nearly impossible to distinguish them. Besides, occlusion is another severe puzzle.

Unlike normal segmentation often utilizing unified threshold for single image, multi-appearance segmentation adopts different thresholds for different objects in the same image. Utilizing the multi-appearance information of small objects, multi-appearance segmentation could automatically vary the threshold in local area and make it more suitable for the small object in certain local area.

\subsubsection{Intention}
Obviously, different thresholds for segmentation product totally different results. As the Fig. \ref{fig_multi_slide} showing, low threshold can not distinguish objects and noisy point well, while too high threshold would lose some objects. In the example of Fig. \ref{fig_multi_slide} we show only three thresholds. In practical situation, the number of layers should be determined according to the specific variance of image. The appropriate threshold is changing for different positions of image. Unified threshold segmentation methods barely embody sufficient discrimination to distinguish false-alarms with real objects.

\begin{figure}[!t]
	\centering
	\includegraphics[width=3.5in]{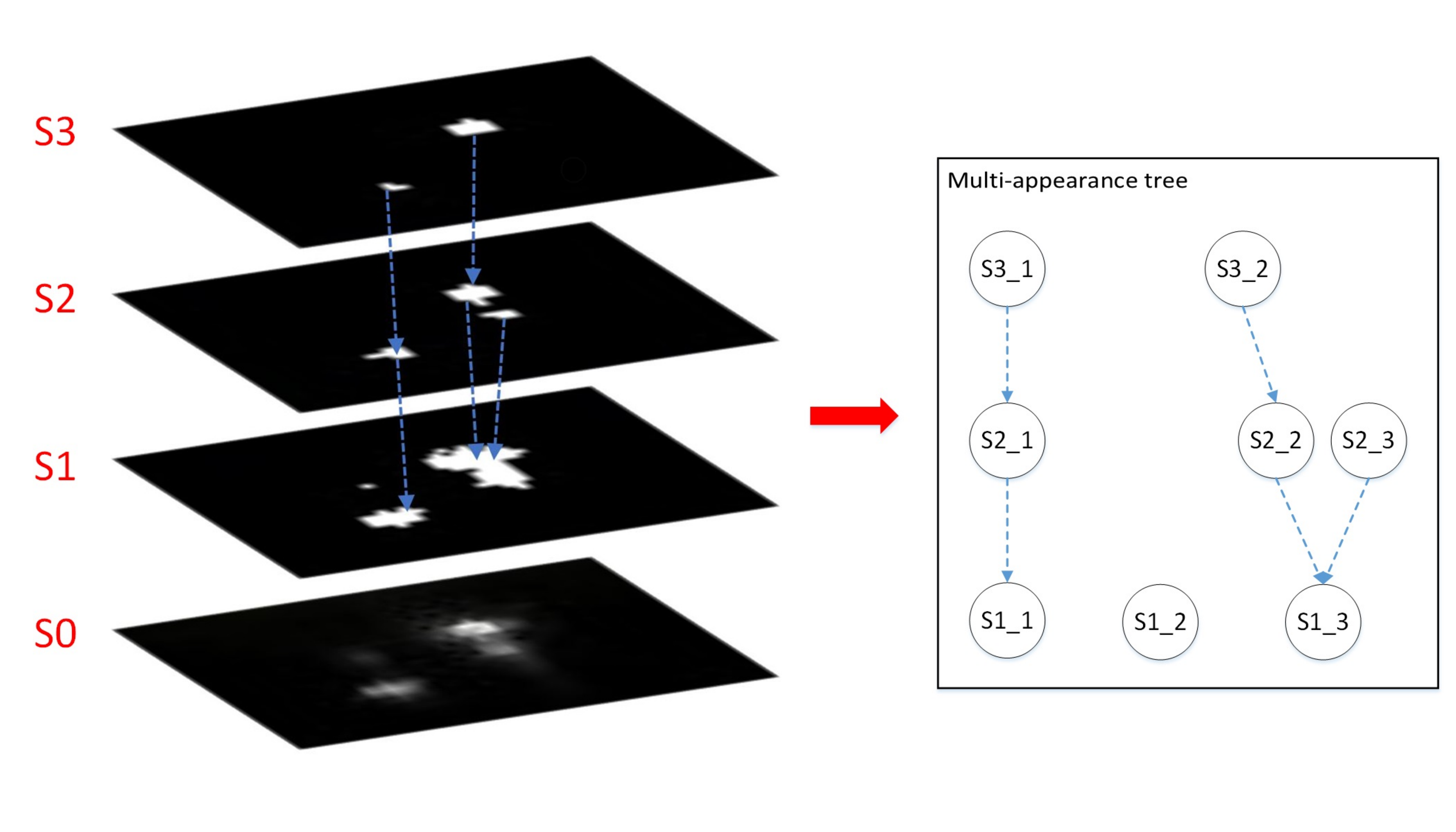}
	\caption{Multi-appearance structure with tree relation. The corresponding tree structure is built according to the relation in binarization slides of original image. As the simple tree example(right side of image) shows, the combination for candidate object selections in the tree with root node $S1\_3$ can only be one from $\{S1\_3\}$,$\{S2\_2,S2\_3\}$ and $\{S3\_2,S2\_3\}$. That is based on the constraint of affiliation relations indicated by the tree structure. You can't select the $S1\_3$ and $S3\_2$ simultaneously, since the thresholds for a local region is unique.}
	\label{fig_multi_slide_tree}
\end{figure}

Each segmentation of gray objects demonstrates its single appearance, which could be interpreted as one slice measurement. From one slice measurement, we can obtain the corresponding object distribution hypothesis score. Sequential layers of slice measurements plus affiliated connections form a small object appearance tree. The affiliated tree of objects in Fig. \ref{fig_multi_slide} is shown as Fig. \ref{fig_multi_slide_tree}. The critical problem in segmentation is to select a more appropriate threshold for each object from all the layers. The criterion for layer selection is to maintain the shape of objects.

\subsubsection{Details of detection method}
Our multi-appearance segmentation is composed of three stages. During the first stage, the gray image is transformed into the corresponding map using \ref{equ_lc_map} and \ref{equ_lc_mean}, then the global threshold binarization is performed on the transformed image. Then, the tree structure is built in second stage based on the information from slides. After that, during the third stage, deep-first-branch-adjustment algorithm is executed for each multi-appearance tree.


\begin{align}
MI_{n}=\frac{1}{N_{u}} \sum_{j=1}^{N_{u}}I_{j}
\label{equ_lc_mean} \\
C_{n} = \dfrac{I_{n}^{2}}{MI_{n}}
\label{equ_lc_map}
\end{align}

In the defined \ref{equ_lc_map}, $MI_{i}$ is the mean intensity of appointed neighbor area. The appointed neighbor area of a certain pixel is the part between outer square windows and inner windows with current pixel at center. The outer and inner width are set to 9 and 3 respectively. $N_{u}$ is the number of pixels in the neighbor area, and $I_{j}$ is the gray level of the $j$th pixel in neighbor area. $I_{n}$ represents the gray level of the central pixel and $C_{n}$ is the final correspond value.

For the first stage, \ref{equ_lc_map} and \ref{equ_lc_mean} is inspired by C.L.P.Chen's work \cite{localcontrastdetTGRS}. However, we remove the minimum search and keep the edges of objects clear. the global threshold, which is the sum of average intensity and $K$ times of variance, is calculated in the beginning. Then, using global threshold as the reference middle value, $n$ layers of thresholds are set with equal interval. For each threshold, a corresponding segmentation is performed, meanwhile acquiring the object information via filling algorithm. The relationship of affiliation between objects from adjacent layers is preserved. After this stage, a multi-appearance tree is built, where a clear tree-like structure is used to describe the relationship between objects from adjacent layers.

\begin{figure}[!h]
	\begin{algorithmic}[1]
		\REQUIRE $ n_{root} $
		\ENSURE $ S_{root} $
		\STATE $S_{child score sum} = 0$
		\STATE Calculate $ S_{appearance} $ according to \ref{equ_s_appearance}.
		\FOR{each $ n_{child} $ satisfying that $ n_{child} $ is child node of $ n_{root} $}
		\STATE $ S_{child score sum} \leftarrow S_{child score sum}+$ Deep first branch adjustment$(n_{child})$
		\ENDFOR
		\IF{$ S_{child score sum} < S_{appearance} $}
		\STATE $ S_{root} \leftarrow S_{child score sum} $
		\ELSE
		\STATE mark $ n_{root} $ as candidate node
		\STATE $ S_{root} \leftarrow S_{appearance} $
		\ENDIF
		
	\end{algorithmic}
	\caption{\textbf{Algorithm}: Deep first branch adjustment}
\end{figure}

Then, Deep-first-branch-flow algorithm is used to mark the candidate nodes in the appearance tree. For children nodes of the candidate node, we just discard that part of tree. Then, the breadth-first search algorithm is used to determinate the final objects, which are the first candidate nodes on the each way of branch starting from the root node to leaf nodes.

We define the appearance score as the product of three score components (intensity, shape, bubble punishment):

\begin{equation}
 S_{appearance}=S_{intensity}S_{shape}S_{bubble punishment} 
\label{equ_s_appearance}
\end{equation}
$ S_{appearance}$ evaluates the object's likelihood based on the consideration that intensity variation will maintain certain stability inside the separated object. In fact, the greatest intensity variation in the image to be detected should be the edges which split the object and background. $ S_{intensity} $ is defined as variance of intensity.
\begin{equation}
S_{intensity}=\dfrac{\sum_{x,y\in O}^{}(I_{x,y}-\bar{I})}{n}
\label{equ_s_intensity}
\end{equation}
We use $ S_{shape} $ to measure the impact of object geometric shape. Accurate segmentation should impel the profile pattern to be glossy and inerratic. $ S_{shape} $ is defined as variance of pixel distance.
\begin{equation}
S_{shape}=\dfrac{\sum_{x,y\in O}^{}(x-\bar{x})(y-\bar{y})}{n}
\label{equ_s_shape}
\end{equation}
And a punishment factor $ S_{bubble punishment} $ is used for the regularization, which treat the non-detected pixel in the object as bubble and employ it to measure the defect as punishment.
\begin{equation}
S_{bubble punishment}=N_{bubble}+1 
\label{equ_s_bubble}
\end{equation}

\section{Our Tracking method for dense small object}
\subsection{Our tracking framework}
We propose a tracking framework with MMHT as hypothesis generator and extended 0-1 program as hypothesis selection component. MMHT is inspired by the TOMHT, using the similar tree hypothesis structure and the idea of limitation for branch expansion. Then extended 0-1 program is proposed and employed to be the substitution of hypothesis enumeration procedure. In the traditional MHT, hypothesis enumeration is carried out for each frame to find the group of hypotheses where each hypothesis is compatible with others. For this intractable problem, we transform it into an extended 0-1 linear program problem, meanwhile implement the many-to-one association assumption through adjunctive binary variables.


The tracking framework is illustrated as the working flow diagram of Fig. \ref{fig_whole_frame}. MMHT produces the most reliable hypotheses of every tree, which are sent to hypothesis selection component as input data. Then, our extended 0-1 program would determine the set of final preserved hypotheses in full consideration of compatible relationship among hypotheses. Noted that our 0-1 program doesn't treat every pair of incompatible tracks in a total hard way. Extended 0-1 program is a soft method with more flexibility to handle some complex circumstances.

\subsection{MMHT to form candidate tracks}
MMHT(management for multi-hypothesis tree) is developed as the generator of potential tracks. We integrate some technologies for hypothesis processing, and design this collection of ordered procedures as our hypothesis management method. MMHT comprises of our hypothesis management method and other plain data processing steps. 

\subsubsection{Tracking description}
Firstly, we introduce the mathematical-expressional form. Assume a sensor scans the surveillance region periodically. The set of measurements received at frame $ t $ is denoted by $ M(t) $:
\begin{equation}
M(t)=\{m_{i}^{t}\}_{N_{t}}^{i=0}, t =1,2,...,N
\label{equ_m_t}
\end{equation}
where $ N $ is the number of frames, $ \{m_{i}^{t}\}_{N_{t}}^{i=0} $ is the $i$th measurement received within frame $t$, and $N_{t}$ is the number of measurements received at frame $t$. In addition, a dummy observation $m_0^t$ is defined for each frame $t$ to denote possible missed detections.

A track hypothesis (we would use hypothesis for short in following content) $T_j^t$ at frame $t$ is defined as a sequence of observations:

\begin{align}
&T_j^t=(m_{j_{1}}^{1},m_{j_{2}}^{2},...,m_{j_{t}}^{t}), m_{j_{n}}^{n}\in M(n)
\label{equ_T_tj}\\
&O_l^t=(T_{1}^{t},T_{2}^{t},...,T_{n_l}^{t})
\label{equ_O_lt}\\
&O^t=(O_1^t,O_2^t,...,O_n^t)
\label{equ_O_t}
\end{align}

This definition constitutes a restriction that one track can contain at most one measurement at a particular frame. Track hypothesis score is associated with each hypothesis to evaluate the likelihood of being the true target. $O_l^t$ is the set of hypothesis tracks, with all the $T_{n}^{t}$ in it possessing same root. $O^t$ is the set of $O_l^t$, representing all tree hypotheses at frame $t$.

\subsubsection{Track management method}

Since our tracking method is designed to be of batch-optimization, 0-1 program will not be executed at every frame, but over a larger span. Before the hypothesis selection in 0-1 program, the hypothesis tree will grow to a colossal scale if no special restriction means is employed.

The principles of new management method are: 1) Hypothesis growth is of more restrictive; 2) The complexity is reduced by various kinds of measures. Our hypothesis management method is flexible and easy to be controlled. This hypothesis management method includes following functions: gating, low-level hypothesis assessment for acceptation or rejection, and hypothesis merging.

\paragraph{Score for moving variability}
Firstly, we defined $S_{MV}$, representing the score of moving variability.
\begin{align}
&S_{MV} (T_j^t)=\dfrac{S_{MV} (T_j^{t-1})(N_C^j-1)+\Delta V_{m_i^t,P_{j_{t-1}}^t }}{N_C^j+1} 
\label{equ_s_mv}\\
&\Delta V_{m_i^t,P_{j_{t-1}}^t }=mahaldist(V_{m_i^t},V_{T_j^{t-1}})
\label{equ_v}
\end{align}
where the following notations are used:
\\$N_C^j:$ depth of hypothesis $j$, which will not exceed the depth of practical hypothesis tree. 
\\$S_{MV}(T_j^t):$ score of leaf hypothesis $j$ at frame $t$ for moving variability;
\\$V_{m_i^t}:$ the velocity of $m_i^t$ at frame $t$ assumed association between $m_i^t$ and $T_j^{t-1}$ is built;
\\$m_i^t:$ measurement $i$ at frame $t$;
\\$P_{j_{t-1}}^t:$ prediction deduced from the $T_j^{t-1}$;
\\$mahaldist(\cdot):$ function to get mahalanobis distance.

The velocity and prediction used here are acquired through correction of Kalman filter. In fact, $S_{MV}(T_j^t)$ can be regarded as an autoregressive model based score, if undo $S_{MV}(T_j^t)$ , we can get following autoregressive formula of $\Delta V_{m_i^l,P_{j_{l-1}}^l }$, with $l$ as the frame number. The initial $S_{MV} (T_j^t)$ is set as zero, so the constant term of this autoregressive formula is zero.
\begin{align}
&S_{MV}(T_j^t )=\sum_{l=1}^{t}a_{l}\Delta V_{m_i^l,P_{j_{l-1}}^l }
\label{equ_s_mvt}\\
&a_l=\frac{2}{N_C+1}(\frac{N_C-1}{N_C+1})^{t-l}
\label{equ_al}
\end{align}
As the result of decreasing weights term used in \ref{equ_s_mvt}, the coefficient $a_l$ decreases as the $l$ descending. The approximate effective order of this autoregressive model varies from 10 to 20 in a correspondingly reasonable short period, depending on the value of $N_C$.

\paragraph{Score test for moving variability}
For each leaf node in hypothesis tree, a $S_{MV}(T_j^t )$ is calculated and maintained via iteration. Then, we use the test equation \ref{equ_aij} to filter out unsatisfying hypotheses before branch growth(or called gating).
$A_{ij}=0$ means the corresponding association won't pass gating.
\begin{align}
&A_{ij}=\begin{cases}
	1&|\Delta V_{m,P}-S_{MV} (T_j)|<th_n,\\
	0& \text{else}.  
	\end{cases}
\label{equ_aij}\\
&th_{n}=\begin{cases}
	(\alpha-N_s)\beta &(\alpha-N_s)>\gamma,\\
	\delta & \text{else}.
	\end{cases}
\label{equ_thn}
\end{align}
\\$th_n:$ multistage threshold for score test;
\\$A_{ij}:$ association filter mask between measurement $i$ and hypothesis $j$, $A_{ij}=1$ indicates permission of association between measurement $i$ and hypothesis $j$;
\\$N_S:$ the number of sustaining frames;
\\$\alpha,\beta,\gamma,\delta:$ parameters for multi-stage threshold.

\begin{figure}[!t]
	\centering
	\includegraphics[width=3.5in]{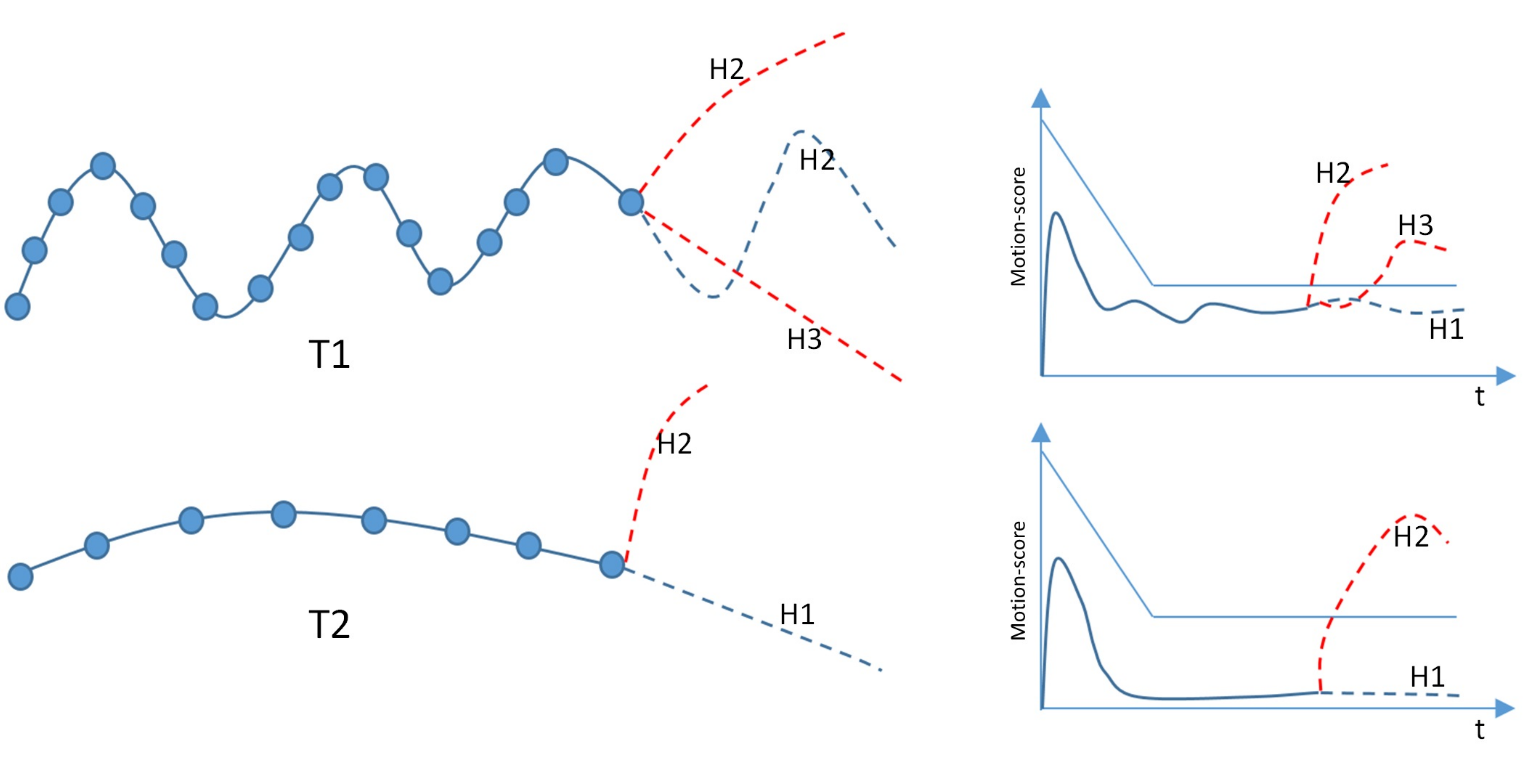}
	\caption{Illustration of score testing. Motion score used here is $|\Delta V_{m,P}$-$S_{MV}|$. The fold line above the score curve indicates our multi-stage threshold $th_n$}
	\label{fig_motion_score}
\end{figure}

We use a figure to illustrate our definition of $S_{MV}$, score test and consideration behind them. As Fig. \ref{fig_motion_score} shows, our score testing is a polygonal line with a descending threshold at first and constant one later.  $|\Delta V_{m,P}-S_{MV} (T_j)|$ should be under the threshold line, once cross then the corresponding hypothesis will be abandoned. We define this score test to describe the strength to maintain the moving pattern of tracks. The moving pattern used here can be interpreted as the degree of volatility. As an autoregressive formulate of $\Delta V_{m,P}$, $S_{MV}$ possesses the attribution of reflecting the expected degree of acceleration. Note that we assume object movement is of constant acceleration. $|\Delta V_{m,P}-S_{MV} (T_j)|$ reflects the deviation of current movement between expected one from the respect of acceleration. Therefore, our score testing is capable to capture the coarse pattern of movement.

For motion score of $T1$ (the curve with frequently changing moving pattern), three potential hypotheses are given. The corresponding diagram of temporary moving score is drawn in the right side. $H1$ and $H2$ is rejected under the result of exploding score, since these two smooth tracks deviate far away for its original fickle pattern. As for $T2$, sudden change of movement also goes against its previous smoothness. This is the assumption we employed here that the volatility of a track will keep in a certain degree within limited time span. To describe this certain degree, we use a simple multistage threshold, in consideration of facilitating the generation of hypotheses by tolerating elusory movement at the beginning.

\paragraph{Hypothesis score}
Indicating the degree of track's likelihood, each hypothesis is assigned with a hypothesis score, which is used for hypothesis selection and merging. The score of hypothesis $j$ at frame $t$ is defined as follows:
\begin{equation}
S(T_j^t)=\omega_{LTM} S_{LTM} (T_j^t)+\omega_{STM} S_{STM} (T_j^t)
\label{equ_track_score}
\end{equation}
$S_{LTM} (T_j^t)$ and $S_{STM} (T_j^t)$ are the scores of hypothesis $j$, in the consideration of long-term motion and short-term motion respectively. $\omega_{LTM}$ and $\omega_{STM}$ are the weights for long-term motion and short-term motion.

$S_{LTM} (T_j^t)$ used here is the original score formulation in traditional MHT. We added $S_{STM} (T_j^t)$ to capture the variability of short term motion, in order to enhance the sensitivity of the score for rapid motion variation. The original score is a slowly changing value, increasing over time for the potential tracks. After a period of updates, it reaches a rather high score and partly losing its sensitivity for motion change. Although some methods like SQRT employed a threshold to detect its variation in high value state, this hard threshold measure can't provide enough agile information for short motion description. So, $S_{LTM} (T_j^t)$ reflects the likelihood of hypothesis from a global view, or in the rather long term.

$S_{LTM} (T_j^t)$ is defined by using cumulative log-likelihood ratio as typical way. 
\begin{equation}
S_{LTM}^t=\sum\Delta S_{LTM}^i (T_j^t)
\label{equ_s_LTM}
\end{equation}

If hypothesis $T_j^{t-1}$ is associated with measurement $m_i^t$ at frame $t$, then the increment of hypothesis score is given by:
\begin{equation}
\Delta S_{LTM}^i(T_j^t)=\begin{cases}
\log (\dfrac{p(m_i^t|T_j^{t-1})P_D}{\lambda_{fa}+\lambda_{nt}})& i\neq 0,\\
\log (1-P_D) & i=0.
\end{cases}
\label{equ_deltas_LTM}
\end{equation}
Where the following notations are used:
\\$m_i^t:$ vector of measurement $i$ at frame $t$;
\\$p(\cdot):$ the probability density function (PDF) of measurement $m_i^t$ conditioned on the one-step prediction of hypothesis $T_j^{t-1}$;
\\$P_D :$ detection probability;
\\$\lambda_{fa}:$ the expected number of false alarms per unit volume of the measurement space per frame(spatial density of clutter);
\\$\lambda_{nt}:$ spatial density of new targets.
\\the initial hypothesis score $\Delta S_{LTM}^i(T_j^t)$ is given as $\log(\frac{\lambda_{nt}}{\lambda_{fa}})$. 

Then, we define $S_{STM}(T_j^t)$ as \ref{equ_STM}.

\begin{align}
&S_{STM}^i(T_j^t)=IC\log (\dfrac{\int_{1}^{IC}t_n-\int_{1}^{IC}|\Delta V_{m,P}-S_{MV} (T_j)|}{\int_{1}^{IC}|\Delta V_{m,P}-S_{MV} (T_j)|})
\label{equ_STM}
\end{align}
$|\Delta V_{m,P}-S_{MV} (T_j)|$ is used for score testing as we mentioned before. This accumulation of deviation could be used to show the deviation tendency according to the known life of a certain object. Then, we use \ref{equ_STM} to acquire long term score based on the history deviation information, where $IC$ is the number of effective emergences. $IC$ counts the occasions when a valid measurement is assigned to the current object. Otherwise, $IC$ decreases as the punishment of missing association. $T_{s}(IC)$ is the accumulation of two-stage threshold $th_n$ used in score testing.   



\paragraph{Hypothesis merging}
The flow diagram of our hypothesis management method is as Fig. \ref{fig_whole_frame}. Then, two stages of merging are carried out to remove logically incompatible hypotheses. The first stage is strong hypotheses merging, which is carried out between different trees. A comparison, between hypotheses with the highest score and highest $IC$, are conducted. These two hypotheses embrace the historic strongest one and the temperate strongest one. Hypotheses would be merged if they share too much measurements in recent period. Looped hypotheses merging is to detect the loop path, where two hypothesis tracks, deriving from the same node, separate on the path and are assigned with the same measurement later. These phenomena may bring exponential growth of hypothesis number if no special treatment is employed.
\begin{figure}[!h]
\begin{algorithmic}[1]
	\REQUIRE $ O^t $
	\ENSURE $ O^t $
	\FOR{each $ O_l^t  \in O^t$}
	\FOR{each $ T_i^t  \in O_l^t \textbf{,} T_j^t  \in O_l^t \textbf{and} T_i^t \neq T_j^t$}
	\IF{$ m_n^t \in T_i^t \textbf{and} m_n^t \in T_j^t $}
	\IF{$ S_{MV}(T_i^t) >= S_{MV}(T_j^t) $}
	\STATE $O_l^t \leftarrow O_l^t - \{T_j^t\}$
	\ELSE
	\STATE $O_l^t \leftarrow O_l^t - \{T_i^t\}$
	\ENDIF
	\ENDIF
	\ENDFOR
	\ENDFOR
\end{algorithmic}
\caption{\textbf{Algorithm}: Loop hypotheses merging(intra-tree)}
\end{figure}

\begin{figure}[!h]
\begin{algorithmic}[1]
	\REQUIRE $ O^t $
	\ENSURE $ O^t $
	\FOR{each $m_n^t \in M(t)$}
	\FOR{each $T_i^t$ satisfying $m_n^t \in T_i^t$}
	\STATE find $ T_{sm}^t $ with max $S(T_{ms}^t)$ .
	\STATE find $ T_{mic}^t $ with max $IC(T_{mic}^t)$ .
	\IF{$T_{mic}^t \neq T_{sm}^t \textbf{and}$ they share more than depth number of detections}
	\IF{$ S(T_{mic}^t) >= S(T_{ms}^t) $}
	\STATE $O_i^t \leftarrow O_i^t - \{T_{ms}^t\}$
	\ELSE
	\STATE $O_j^t \leftarrow O_j^t - \{T_{mic}^t\}$
	\ENDIF
	\ENDIF
	\ENDFOR
	\ENDFOR
\end{algorithmic}
\caption{\textbf{Algorithm}: Strong hypotheses merging(inter-tree)}
\end{figure}

For inter-tree hypothesis merging, hypothesis score is employed as assessment criteria. As to intra-tree merging, temporary score is used.


\paragraph{Birth and death of hypothesis}
Those detections that can't link to any hypothesis would be regarded as the starts of new objects. We employ SQRT test \cite{ReidMHTTAC} to decide whether or not to delete a hypothesis that has no linked detection to update. 

We maintain a score rank for each hypothesis tree. Once a hypothesis is terminated, it will be added to the corresponding position of the rank according to its score. Besides, all existing hypotheses will be added to the rank at the last frame of each batch. For final extended 0-1 program based intra-tree hypothesis selection, we provide only top 20 percent of hypotheses as input, which will release most hypotheses with less possibility and reduce computational complexity burden.  

The main function of hypothesis management method in our paper is to enhance the quality and lower the quantity of hypothesis.

\subsection{Overall hypothesis selection as an extended 0-1 program}
The more objects appear in the scene, the more frequently occlusions would happen. In fact, occlusion problem still is the most momentous challenge.


Various occlusion situations significantly aggravate the complexity of data association. In our observations, the most knotty and deceptive situation is the one with objects moving proximately. Proximate motion means they have similar speed while they encounter each other. Occlusion will last for a longer period with tremendous probability to lose the tracks.

\subsubsection{One-to-many association for intractable occlusions}
When we focus on the approximative motion occlusion, utilizing one-to-many association may be more adapted than one-to-one association.

\begin{figure}[!t]
	\centering
	\includegraphics[width=3.3in]{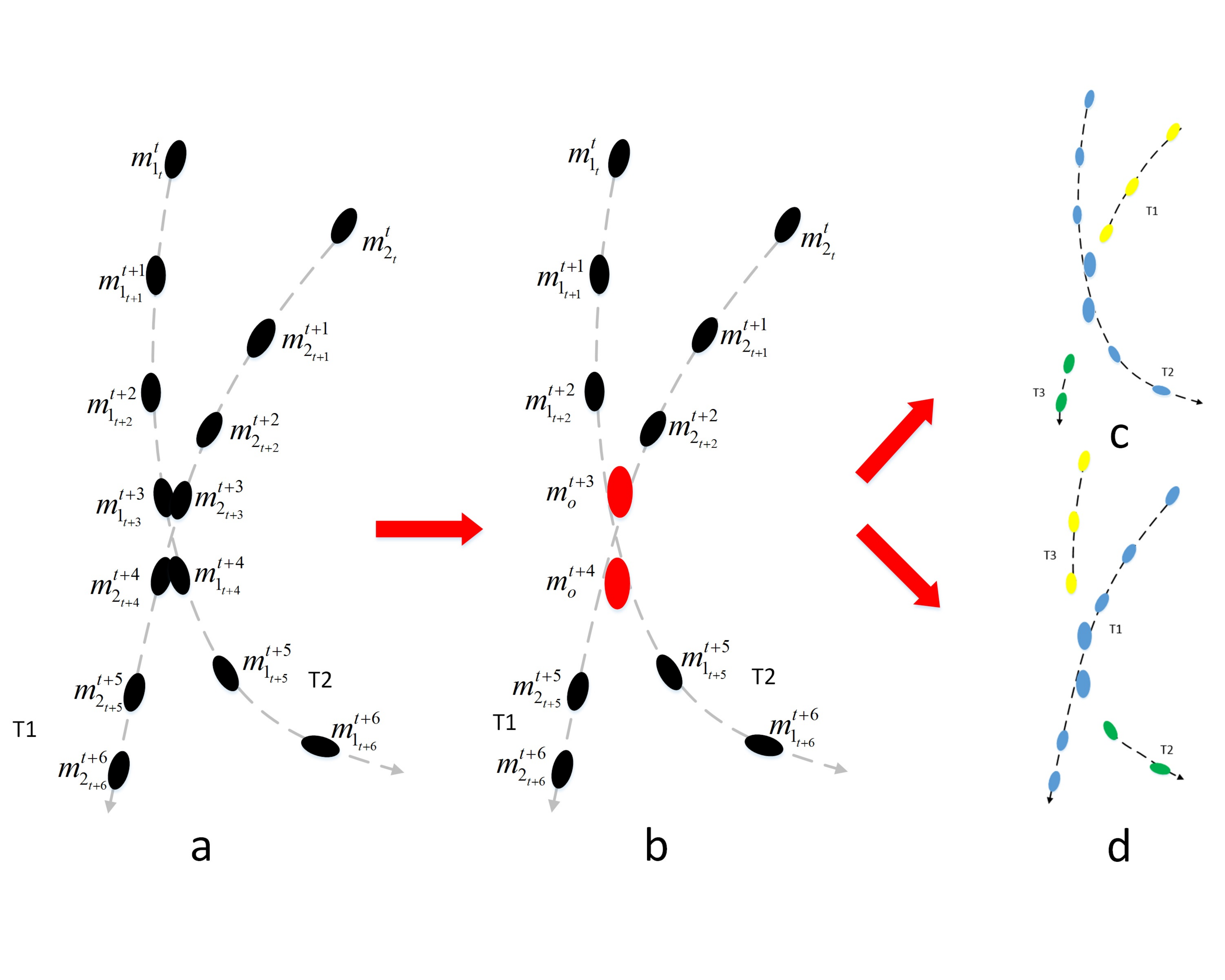}
	\caption{Two tracks with overlapped detections in crossing period are illustrated as sub-figure \textbf{a}. As a result of intricacy of the mixture objects, those overlapped detections are likely to be recognized as a single detection while some are missed or fused. Hence, sub-figure \textbf{b} is the real input data for tracking method, where $m_{1_{t+3}}^{t+3}$ and $m_{2_{t+3}}^{t+3}$ are regarded as single detection  $m_{o}^{t+3}$, same as $m_{o}^{t+4}$. As for one-to-one association, the ownership of certain detection is unique, which gives rise to the scramble for key detections between tracks. The results of that may be like the sub-figure \textbf{c} and \textbf{d} presenting. Some hypotheses lose and can't maintain the continuity of their trajectory or even being totally denied due to the missing of ownership for the key detections.}
	\label{fig_onetomany}
\end{figure}

Most of past studies assumed that at most one object is associated with each measurement. As for the rock-ribbed occlusion with approximative motion, during the occlusion period, two or more objects are detected as only one measurement. As Fig. \ref{fig_onetomany} illustrates, if you insist the assumption of one-to-one association, some of the tracks without corresponding associated measurements will emerge a great gap, which would influence the correct formation of tracks. On the contrary, one-to-many association facilitates the formation of each hypothesis through the one measurement detected zone.

\subsubsection{Formula for one-to-many association}
Track selection is naturally a binary linear program problem in the view of treating each selection of hypothesis as a binary switch of variable. The classical formula of one-to-one association (such as \cite{01programTAC}) is like following.

\begin{gather}
\min C=\min \sum_{n}\xi_{T_{n}}C_{T_n}\\
s.t. \ \xi_{T_{i}}\xi_{T_{j}} = 0, \quad T_{i} \cap T_{j} \ne \emptyset\ and \ i\neq j \label{st_c_1}\\
\xi_{T_n} \in \{0,1\} \\
\label{equ_min_c}
\end{gather}

$C_{T_n}$ is defined as the cost of hypothesis $n$. $\xi_{T_{i}}$ is a binary representation of $T_{i}$. To make it feasible and efficient to take into account the degree of mutual exclusion, we redesign an extended 0-1 linear program with adjunctive binary variable $I_{ij}$. The hypothesis selection is as following formula.

\begin{gather}
\min C=\min \sum_{n}\xi_{T_{n}}C_{T_n} + \sum_{i,j}I_{ij}C_{I_{ij}}\\
s.t. \quad I_{ij} \geq  \xi_{T_i}+\xi_{T_j}-1 \label{st_n_1}\\
I_{ij} \leq (\xi_{T_i}+\xi_{T_j})/2 \label{st_n_2}\\
I_{ij},\xi_{T_{n}} \in \{0,1\}
\end{gather}

$I_{ij}$ represents the index of incompatibility between hypothesis $i$ and $j$. $I_{ij}$ will equal to 1 while hypothesis $i$ and $j$ own same measurement. $I_{ij}=0$ indicates that hypothesis $i$ and $j$ are totally irrelevant with each other. The constraint \ref{st_n_1} and \ref{st_n_2} ensure that while $T_i$ and $T_j$ are both selected, $I_{ij}$ will be 1, otherwise $I_{ij}$ equals to 0. $C_{I_{ij}}$ represents the cost of incompatibility to select both hypothesis $i$ and $j$.

Unlike the classical formula, where a simple constrain is employed to restrict collision, we utilize adjunctive binary variables $I_{ij}$ to take in charge of the considerations for mutual exclusion and one-to-many assignment. In fact, the new term $I_{ij}C_{I_{ij}}$ is analogous to $\xi_{T_{i}}\xi_{T_{j}}C_{I_{ij}}$, which makes the program a quadratic problem as well as an intractable NP-hard problem.

Without the adjunctive binary variables $I_{ij}$, the problem would turn into a quadratic one for the sake of necessary introduction of $\xi_{T_{i}}\xi_{T_{j}}C_{I_{ij}}$. It is a great computation mitigation to solve a linear program problem instead of a quadratic one. Besides, by the introduction of $I_{ij}$ and $C_{I_{ij}}$, the restriction for incompatible hypotheses is steerable. We slightly alleviate such restriction to encourage robust long hypothesis formation. 

\subsubsection{Details of extended 0-1 linear program}
In fact, those adjunctive binary variables $I_{ij}$ with incompatible $i,j$ couples take a quite small portion. So, the total number of variables maintains a limited scale, which is tractable and actually pretty fast as our experiments show.

In consideration of characters of featureless small object, we proposed following criteria as cost of hypothesis. Motion would be the critical component in it.
\begin{equation}
	C_{T_n} = -K\log \frac{N_s^n}{L_t^n+1}
	\label{equ_ctn}
\end{equation}
Where $N_s^n$ is the number of sustaining frames, and $L_t^n$ is the score of leaf hypothesis $T_n$ at frame $t$. $K$ is used as adjustment coefficient. As for the cost of incompatible hypotheses couple, we use following formula.
\begin{equation}
	C_{I_{ij}}=\begin{cases}
	N_I^{ij}C_{IN} &N_I^{ij}<L_T,\\
	\infty & \text{else}.
	\end{cases}
\end{equation}
$N_I^{ij}$ is denoted to be the number of incompatible measurements between $T_i$ and $T_j$. $C_{IN}$ is a constant, representing the cost multiplier for incompatibility of single measurement sharing. $L_T$ is settled as the threshold of tolerable upper limit. If $N_I^{ij}$ exceeds, an outrageous value will be assigned to $C_{I_{ij}}$, indicating that $T_i$ and $T_j$ are totally antipathetic to each other.

Then, after the building of 0-1 program for each batch of frames, we use lpsolve to solve this problem. Lpsolve is reasonably efficient. Thanks to the rational assumptions and optimized constraints, solving progress spends totally affordable time. Detailed time consumption and analyses will be presented in the next section.

\section{Experiments}
In this section, we describe several experiments to verify the performance of our segmentation and tracker. The detailed data and contrastive analysis will be presented to prove the capacity of proposed methods.
\subsection{Comparing methods}
Firstly, those methods used for comparison and its parameter settings shall be listed. Two parts of experiments were performed, including different segmentation and tracking methods respectively. 
\subsubsection{Segmentation methods}
\paragraph{LCM based detection}
Using contrast based template operator would help to extract salient points, and also will be conducive to local optimization \cite{localcontrastdetTGRS}. Its sensibility to small object in the low-SNR circumstance really impresses us. For simplification of expression, we use LCM hereinafter to indicate the LCM based detection method.
\paragraph{Top-hat filter}
Top-hat operation can extract small elements and details from image, which had been found truly useful for small object detection.
\subsubsection{Tracking methods}
\paragraph{MHT}
The parameters for cox's implementation \cite{COXMHTPAMI} of MHT mainly derived from default setting. Only few changes were made to make it more adaptive for some scenarios. We also compared the speed of cox's with our method when running on the same platform and implementing in the same language (C++).
\paragraph{SGTS}
We also employ SGTS as another compared method. It's implemented in MATLAB, so no comparison in speed about SGTS would be presented. The number of semi-greedy solutions generated before selection was set at 60. 
\paragraph{MDA}
The duality gap of the termination criterion was set to 0.02. We defined the maximum number of iterations to 100. The Lagrange multiplier updating scheme applied was the heuristic price update, because it's believed to be more efficient and can fully exploit the structure of the intermediate feasible solutions found by the Auction algorithm.
\paragraph{GRASP-MHT}
Greedy randomized adaptive search was applied in multi-object tracking by Murphey \textit{et.al} \cite{Murphey2010A} and Robertson \textit{et.al} \cite{Robertson2001A}. We used a MATLAB implementation from ren's \cite{GRASPMHTTAES}, where GRASP is used as an engine of hypothesis generation in the MWISP formulated TOMHT \cite{GRASPMHTTAES}. Parameters $nv$, $np$, and $nitr$, which were used to control the amount of computation in candidate construction, were set to 20, 20, and 3, respectively. In the hypothesis pruning procedure, those with a probability lower than $10^{-6}$ are discarded. 

\subsection{Performance metrics}

\subsubsection{Detection Metrics}
\paragraph{Detection Rate(\textbf{DR}$\uparrow$)}
We use the definition that $DR=NC/NT$, where $NC$ and $NT$ are the number of correctly detected objects and true objects respectively. 
\paragraph{False Alarms(\textbf{FA}$\downarrow$)}
We have $FA=NIC/N$ while $NIC$ is the number of incorrectly detected objects. Meanwhile, $N$ is the length of the sequence.
\paragraph{Standard Deviation of Detection Rate(\textbf{DR-STD}$\downarrow$)}
Lower standard deviation of detection rate indicates better stability of detection method.

\subsubsection{Tracking Metrics to Treat Object as Point}

\paragraph{Optimal Sub-pattern Assignment Distance(\textbf{OSPA-T}$\downarrow$)}
The modified Optimal Sub-pattern Assignment Distance proposed by Branko Ristic is widely used as measure index in multi-object tracking. $c$ and $\ell$ are set to 25 as default, and p is 2.

\paragraph{Track Completeness Factor(\textbf{TCF}$\uparrow$)}
Track Completeness Factor measures how well we detect a given object after the association\cite{TFTCF}. $tol$ used in \textbf{TCF} is set as 15.
\paragraph{Track Fragmentation(\textbf{TF}$\uparrow$)}
Track Fragmentation measures how well we maintain identity\cite{TFTCF}. $tol$ used in \textbf{TF} is set as 15 too.

\subsubsection{Tracking Metrics Assuming that Objects Occupy Certain Space}
We used some metrics based on the reference from visual multi-object tracking like pedestrian tracking, which attracts great attention recently. Those metrics were designed for object with a certain size and detection box, rather than a small object with few pixels. However, the small objects in our dense tracking scenario show up more than just few pixels and occupy considerable space(they are still small object with not more than one hundred pixels). Higher density actually exaggerates the effect of their size. Those metrics were designed to evaluate the complex scenario with massive occlusions, which is more complicated than traditional scenario. Applying those new performance metrics could augment diversity and reference value of our result. Under such consideration, we utilized the CLEAR MOT metrics \cite{MOTChallenge2015}.

\paragraph{Number of Identity Switch(\textbf{IDSW}$\downarrow$)}
Identity Switch counts the number of emergences when a ground truth target $i$ is matched to hypothesis $j$ and the last known assignment was $k$($k \neq j$) \cite{MOTChallenge2015}. 
\paragraph{Multiple Object Tracking Accuracy(\textbf{MOTA}$\uparrow$)}
Thanks to its expressiveness, the Multiple Object Tracking Accuracy \cite{MOTChallenge2015} may be the most widely used figure in evaluating a tracker's performance. The definition of Multiple Object Tracking Accuracy is as \ref{equ_mota}:
\begin{equation}
MOTA=1-\dfrac{\sum_t(FN_t+FP_t+IDSW_t)}{\sum_tGT_t}
\label{equ_mota}
\end{equation}
It combines three different sources of errors, where $t$ is the index of frame and $GT$ is the number of ground truths. $FN$ is the number of false negatives, and $FP$ is the number of false positives. 
\paragraph{Multiple Object Tracking Precision(\textbf{MOTP}$\uparrow$)}
The Multiple Object Tracking Precision is the average dissimilarity between all true positives and their corresponding ground truth targets \cite{MOTChallenge2015}.

\paragraph{Ratio Misses Over Total Number(\textbf{FN$\downarrow$})}
The ratio misses in the sequences over the total number of objects presenting in all frames \cite{TFTCF}.
\paragraph{Ratio False Positive Over Total Number(\textbf{FP}$\downarrow$)}
The ratio False Positives over the total number of objects presenting in all frames \cite{TFTCF}.

\paragraph{Recall(\textbf{REC}$\uparrow$)}
The number of correctly matched detections divided by the total number of detections in ground truth.
\paragraph{Precision(\textbf{PRE}$\uparrow$)}
The number of correctly matched detections divided by the total number of output detections.

We use up arrow $\uparrow$ to represent that higher score indicates better result.The opposite of that, down arrow $\downarrow$, means preference to lower score.

\subsection{Dataset}
Two datasets were used in out experiments, denoted as $Larva$ and $Verti\_Hat$ respectively.
\paragraph{$Larva$}
Scene with movements of micro-animal is our first-line application for this paper, three segments of video data were used in this paper. All of them contain nearly one hundred objects in a single frame, meanwhile some appear frequent occlusions. 

They were captured in different conditions with differentiated image qualities. Three sequences represent three degrees of difficulty, the image quality of $Larva\_s2$ is a bit lower than $Larva\_s1$, due to extra ripple interferes deriving from sensor noise. In $Larva\_s3$, focal length is changed several time, resulting in drift of the focus plane. Objects could suddenly become blurry in couple frames and be missed by detector.

We sampled the video images at regular interval, and took half frames for experiments to reduce the computational complexity. The ground truth was produced based on the tracking results of few tracking methods enumerated before. Unlike experiments, the video used for pre-designation is full-frame without down sampling. So it'd be of less ambiguity because that provides more data to fill the uncertainty gap. Then, we checked and corrected the trajectories manually used those results as reference to reduce efforts, especially focusing on key frames where occlusions or false alarms happen.

\paragraph{$Verti\_Hat$}
This pedestrian video was from the vertical view, where around 60 persons with hat walk round an appointed region. After adding disturbance and removing some measurements, this scene becomes of considerable difficulty. This dataset comes from Ren's work \cite{GRASPMHTTAES}, we intercepted the first 600 frames of the pedestrian video, then down sampled it at 5Hz as the author did. Finally, 120 frames of images were used in this scenario test, providing twice length of used data in Ren's paper.

\subsection{Parameters}
The depth of hypothesis tree was set to 6, parameters for multi-stage threshold were set as $\alpha=20, \beta=0.8, \gamma=10, \delta=6$. $\omega_{LTM}$ and $\omega_{STM}$ were both 1. As for parameters in $S_{STM}$,$P_D = 0.9, \lambda_{fa} = 1e^{-6},\lambda_{nt} =1e^{-8}$. $K,C_{IN},L_T$ in the programming part were set to 5,1,5 respectively. Optimization batch length was set to 20 with a consideration window of 40, which is the length of windows containing former frame. Any frame in it would be took into account when track selection is underway.

Usually, if the distance between a ground truth and a detected position is within a threshold, then the detection is declared as being correct \cite{Kim2012Scale}. In consideration of our tracking scenario with dense and bigger objects, moreover the resolution of sequence used here is more than 3 hundred thousand compared to about 60 thousand in C.L.P.Chen's\cite{localcontrastdetTGRS}, we chose 15 as a distance threshold according to proximate increasing of proportion. For the metrics assuming that objects occupy certain space, we treat each object as in a box with an radius of 15.

\subsection{Experiment arrangement}
Two components of experiments were arranged and performed. They focus on segmentation method and tracking method respectively.
\subsubsection{Segmentation method}
We arranged a detection experiment to test our multi-appearance segmentation compared with classical one (Top-hat) and state-of-art one (LCM). 



\begin{table*}[!t]
\renewcommand{\arraystretch}{1.3}
\caption{Performance Comparison Between Our Segmentation Method and Others for the $Larva$ Dataset}
\label{table_segmentaion}
\centering
\begin{tabular}{c|c c c c c c c}
\hline
\hline
\textbf{Dataset} & \textbf{Method} & \textbf{DR} & \textbf{DR-STD} & \textbf{FA} & \textbf{OSPA-T} & \textbf{MOTA} & \textbf{MOTP}\\
\hline
\multirow{2}{*}{$Larva\_s1$}
& MAS & 90\% & 0.0473 & 29 & 13.5 & 67.3 & 90.3 \\
& Top-Hat & 79.3\% & 0.0524 & 24.9 & 14.6 & 48.7 & 83.5\\
& LCM & 84.6\% & 0.0653 & 13.5 & 16.8 & 52.6 & 80.9\\
\hline
\hline
\end{tabular}
\end{table*}

First of all, we used F-score to pre-evaluate the three detection methods. F-score is widely used for the assessment with definition of $ F=2*recall*precision/(recall+precision) $, which takes both recall and precision into account. A best K, which is acquired by traversal on samples of video fragment(part of $Larva$ dataset), was chosen for every segmentation according to the F-score curve. Then, we got best $ K= 1.8,3.5,5.8$ for LCM,MAS,Top-hat respectively. In fact, $ K\in[3,5] $ recommended in \cite{localcontrastdetTGRS} provides poor results in our dense object scenario, with DR less than seventy percent. Dense objects occupy more space. So a befitting $K$ should be determined through sample tests.

\begin{figure}[!t]
	\centering
	\includegraphics[width=3.2in]{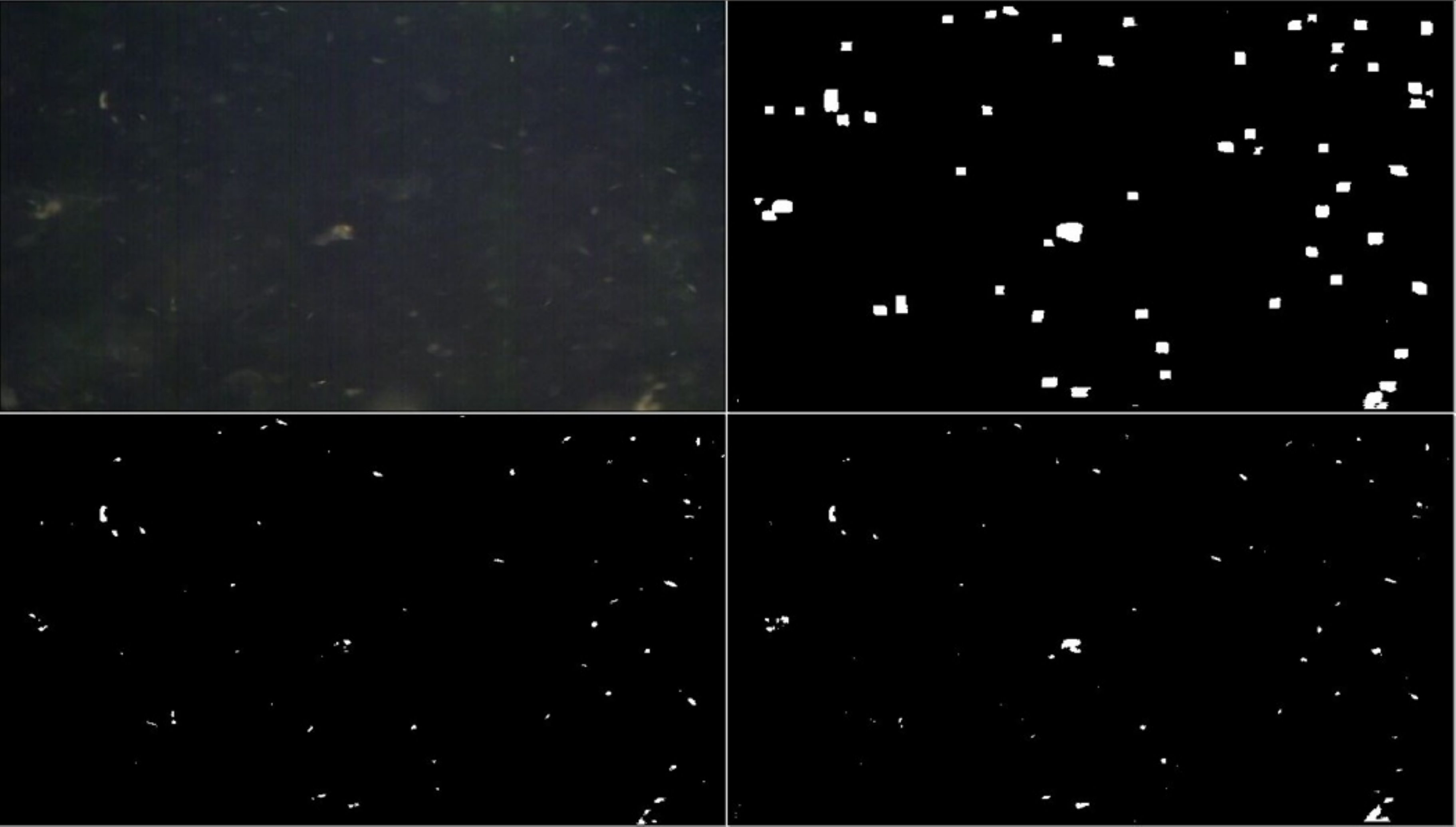}
	\caption{Segmentation results of three methods for frame 14 of $Larva\_s1$. Upper left picture is the original image, upper right,lower left and lower right pictures are outputs of LCM, Top-hat and MAS respectively}
	\label{fig_seg_4pics}
\end{figure}

Then, we used the best $K$ for each detection method to ensure that each segmentation yields their best result. Several indexes are listed in table \ref{table_segmentaion}, including some tracking indexes to assess detection from the view of the whole tracking procedure. In fact, from the F-score, our segmentation is not superior to LCM, however, LCM exposes some problems in final result. As Fig. \ref{fig_seg_4pics} show, expanding objects go against the precise segmentation in the response map of LCM. We also can discover the impact of this defect in final tracking metrics. That is why we developed our new segmentation to improve the results of tracking. Besides, in consideration of our main purpose of segmentation (i.e integrating segmentation into tracking), evaluating the finally tracking consequences would be more objective and revealing. The tracking procedure used for three detection methods is totally identical with the same settings.


Table \ref{table_segmentaion} shows that our method achieves the best detection rate and lowest DR-STD. Meanwhile, it yields more false alarms. The priority mission of detection to manifest objects, secondly to delete false alarms. According to the final tracking result, we detection method facilitates the growth of MOTA \cite{MOTAStiefelhagen2006} by more than ten percent. For other tracking indexes, $MAS$ get best scores too. The results demonstrate the detection capability of $MAS$ and excellent cooperation between our detection and tracking method.

Our detection method is designed to detect dense object and cooperate with our tracking method. So it may not be appropriate for general detection of small object. MAS takes better consideration of occlusion and crossing situations for dense object, and utilizes the multi-appearance tree to retain distinguishability, discouraging tiny object from covering of bigger agglomerate object. LCM uses max-pooling like operation to enhance and agglomerate object, which could be beneficial to remove false alarms. However, agglomerate object may deteriorate the situation of occlusion and degenerate distinguishability.

\subsubsection{Tracking method}
We present a series of experiments here, with purposes of testing overall performance and one-to-many assumption. Apart from comparison experiments with other tracking methods, we try to explore and figure out the effect of our new idea employed in this paper.

Four tracking methods listed in last subsection were used in this comparison. Two datasets used in this part indicate traditional scenario and dense scenario, denoted as $Verti\_Hat$,$Larva$ respectively.

\paragraph{Scenario $Verti\_Hat$}

The results are presented in Table \ref{table_tracking_hat}. Our tracking method outperforms others in the principal metrics such as OSPA-T, MOTA and IDSW. OSPA-T is a critical metrics in the traditional tracking scenario. Our tracking method achieves near 20\% decreasing in this distance compared with the second method,i.e GRASP-MHT. Besides, lower IDSW is a remarkable feature of our method, which can be easily noticed in the following experiment results as well.

\paragraph{Scenario $Larva$}
The primary problem of this paper is to develop an efficient method for dense object in the clutter background. Larva tracking from biology applications is such very situation. We collected three sequences with different qualities of image to represent different tracking conditions.

The final row in Table \ref{table_tracking_larva} is the weighted average of metrics from three sequences, where the weight is the corresponding frame number divided by the total frame number. Note that IDSW,FN and FP are different, they were sums of three sequences. For this aiming scenario, our method outperforms other methods and ranks first in 6 items of total 10 metrics. The recall figure of our method is at an ordinary level, since we try much effort to reduce the computational complexity and also bring in sacrifices of some valid hypotheses. After the trade-off of complexity and performance, we reach a balance to keep the computation totally tractable and even super fast in this scenario, with decent results preserved at the same time. 

\begin{figure}[!t]
	\centering
	\subfloat[Ground truth of $hat$]{\includegraphics[width=1.5in]{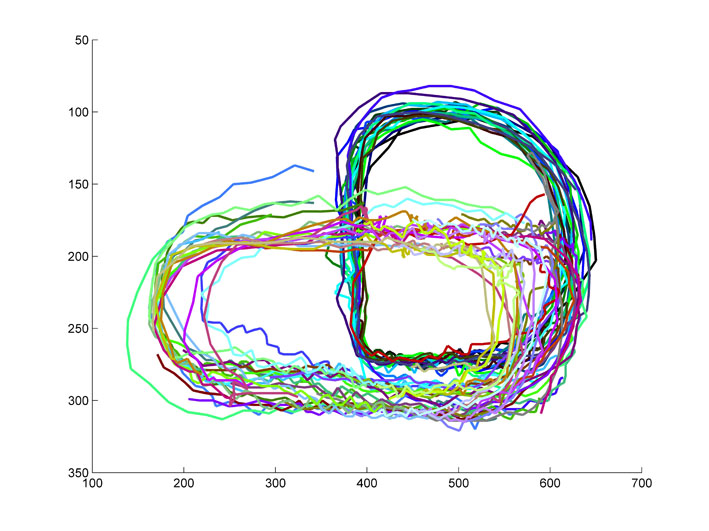}%
		\label{fig_hat_gt}}
	\hfil
	\subfloat[Ground truth of $larva\_s1$]{\includegraphics[width=1.5in]{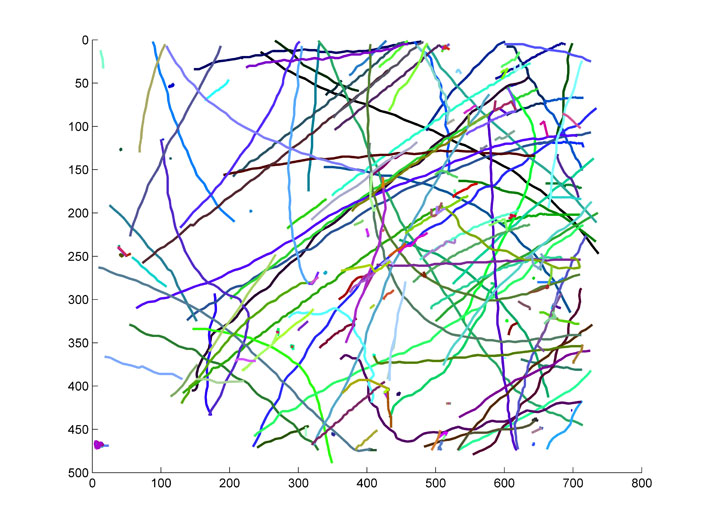}%
		\label{fig_larva_s1_gt}}
	\hfil
	\subfloat[Ground truth of $larva\_s2$]{\includegraphics[width=1.5in]{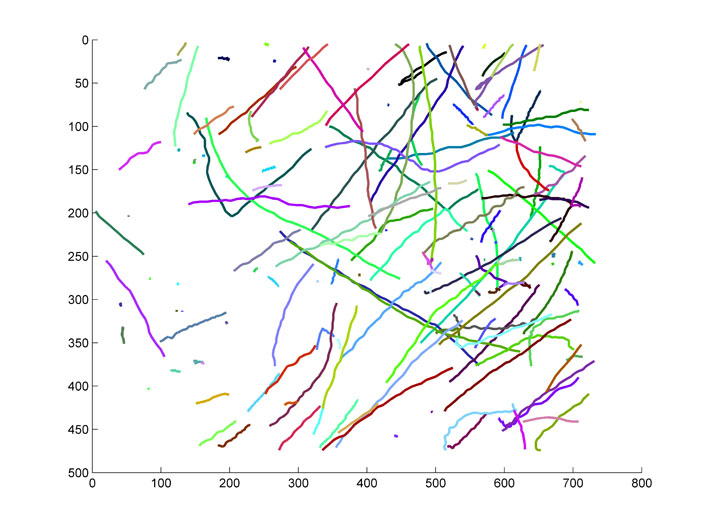}%
		\label{fig_larva_s2_gt}}
	\hfil
	\subfloat[Ground truth of $larva\_s3$]{\includegraphics[width=1.5in]{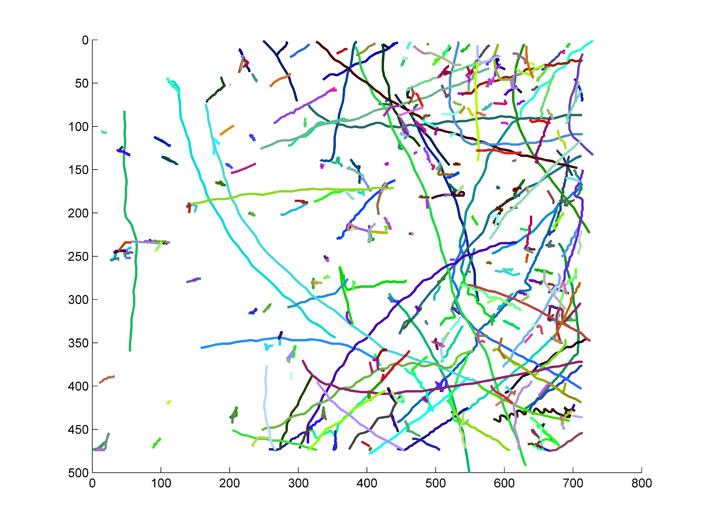}%
		\label{fig_larva_s3_gt}}
	\caption{Ground true of datasets.}
	\label{fig_dataset_gt}
\end{figure}

\begin{table*}[!t]
\renewcommand{\arraystretch}{1.3}
\caption{Performance Comparison Between Our Tracking Method and Others Methods for the $Verti\_Hat$ Dataset}
\label{table_tracking_hat}
\centering
\begin{tabular}{c|c c c|c c c c c c c}
	\hline
	\hline
	\multirow{2}{*}{\textbf{Method}} &
	\multicolumn{3}{c|}{\textbf{Traditional Metrics}} &
	\multicolumn{7}{c}{\textbf{CLEAR MOT Metrics}}\\
	\cline{2-4}
	\cline{5-11}
	& \textbf{OSPA-T} & \textbf{TF} & \textbf{TCF} & \textbf{MOTA} & \textbf{MOTP} & \textbf{IDSW} & \textbf{FN} & \textbf{FP} & \textbf{REC} & \textbf{PRC}  \\
	\hline
	Ours & 14.7 &2.1 & 0.8 & 56.9 &	71.3 &	68 &	2026 &	1011 & 71.9 & 83.7\\
	Cox's MHT & 21.9 & 12.2 & 0.9 & 50.6 & 71.4 & 711 & 1653 & 1196 & 77.0 & 82.3\\
	GRASP-MHT& 18.2 & 2.2 & 0.6 & 52.0 & 71.2 & 302 & 2009 & 1148 & 72.1 & 81.9\\
	MDA & 20.4 & 4.8 & 0.6 & 45.3 & 71.0 & 514 & 2334 & 1094 & 67.6 & 81.6\\
	SGTS & 20 & 3.5 & 0.5 & 46.3 & 70.9 & 457 & 2251 & 1156 & 68.7 & 81.1\\
	\hline
	\hline
\end{tabular}
\end{table*}

\begin{table*}[!t]
	\renewcommand{\arraystretch}{1.3}
	\caption{Performance Comparison Between Our Tracking Method and Others Methods for the $Larva$ Dataset}
	\label{table_tracking_larva}
	\centering
	\begin{tabular}{c|c|c c c|c c c c c c c c c}
		\hline
		\hline
		\multirow{2}{*}{\textbf{Dataset}} &
		\multirow{2}{*}{\textbf{Method}} &
		\multicolumn{3}{c|}{\textbf{Traditional Metrics}} &
		\multicolumn{5}{c}{\textbf{CLEAR MOT Metrics}} \\
		\cline{3-5}
		\cline{6-12}
		& & \textbf{OSPA-T} & \textbf{TF} & \textbf{TCF} & \textbf{MOTA} & \textbf{MOTP} & \textbf{IDSW} & \textbf{FN} & \textbf{FP} & \textbf{REC} & \textbf{PRC}  \\
		\hline
		\multirow{5}{*}{\textbf{Larva\_s1} (typical scene)} &
		Ours & 13.4 & 1.2 & 0.7 & 67.6 &  90.0 & 44 & 1939 & 680 & 76.4 & 90.2\\
		& Cox's MHT & 15.7&	3.2&	0.8&	48.8&	87.1&	282&	1080&	2847 & 86.9 & 71.5\\
		& GRASP-MHT & 17.6&1.7&	0.6&	57.3&	85.6&425&	1641&	1450 & 80.1  &82.0\\
		& MDA  & 17.2&	2.2&	0.6&	56.7&	86&	431&	2094&	1035&74.5  &85.6\\
		& SGTS & 17.3&	1.9&	0.6&	56.1&	85.7&	533&	1657&	1422& 79.9 & 82.2\\
		\hline
		\multirow{5}{*}{\textbf{Larva\_s2} (numerous objects)} &
		Ours &12.7 & 1.0 &  0.8 &  57.0 & 93.0 & 16 &	943 &	831 & 77.3 & 79.5\\
		& Cox's MHT & 16.8&	2.4&	0.9&	28.5&	88.6&	164&	478&	2331&88.5  &61.2\\
		& GRASP-MHT & 17.3&	1.3&	0.7&	33.5&	85.3&	243&	860&	1657&79.3  &66.5\\
		& MDA  & 17.2&	1.8&	0.7&	37.5&	86.4&	251&	1098&	1244&73.5  &71.0\\
		& SGTS & 17.7&	1.4&	0.6&	31.5&	85.4&	280&	946&	1614&77.2  &66.5\\
		\hline
		\multirow{5}{*}{\textbf{Larva\_s3} (focus drift)} &
		Ours &17.6 & 1.1 & 0.6 & 50.6 & 89.3 & 52 &	4790 &	1051 & 59.9 & 87.2\\
		& Cox's MHT & 17.4&	2.2&	0.7&	46.4&	86.8&	313&	3763&	2323&68.5  &77.9\\
		& GRASP-MHT & 18.8&	1.3&	0.5&	43.2&	84.5&	393&	4750&	1638&60.2  &81.4\\
		& MDA  & 19.1&	1.9&	0.5&	42.1&	85.1&	432&	5193&	1283&56.5  &84.0\\
		& SGTS & 19.1&	1.7&	0.5&	42.1&	84.6&	500&	4825&	1589&59.6  &81.7\\
		\hline
		\multirow{5}{*}{\textbf{Larva} (average)} &
		Ours & 15.1&	1.2&	0.6&	58.8&	90.1&	112&	7672&	2562&	69.5&	87.4\\
		& Cox's MHT & 16.6 & 2.7 & 0.7 & 44.9 & 87.2 & 759 & 5321 & 7501 & 79.2 & 72.8\\
		& GRASP-MHT & 18.1 & 1.5 & 0.5 & 47.9 & 85.1 & 1061 & 7251 & 4745 & 71.5 & 79.5\\
		& MDA  & 18 & 2 & 0.5 & 47.7 & 85.7 & 1114 & 8385 & 3562 & 66.6 & 82.8\\
		& SGTS & 18.1 & 1.8 & 0.5 & 46.6 & 85.2 & 1313 & 7428 & 4625 & 70.8 & 79.7\\
		\hline
		\hline
	\end{tabular}
\end{table*}

\begin{table*}[!t]
	\renewcommand{\arraystretch}{1.3}
	\caption{Performance Comparison Between Our Tracking Method and Its Variation Under One-to-one Constraint for the $Larva\_s2$ Dataset}
	\label{table_tracking_larva_s1_one_to_one}
	\centering
	\begin{tabular}{c|c c c|c c c c c c c}
		\hline
		\hline
		\multirow{2}{*}{\textbf{Method}} &
		\multicolumn{3}{c|}{\textbf{Traditional Metrics}} &
		\multicolumn{7}{c}{\textbf{CLEAR MOT Metrics}}\\
		\cline{2-4}
		\cline{5-11}
		& \textbf{OSPA-T} & \textbf{TF} & \textbf{TCF} & \textbf{MOTA} & \textbf{MOTP} & \textbf{IDSW} & \textbf{FN} & \textbf{FP} & \textbf{REC} & \textbf{PRC}  \\
		\hline
		one-to-many & 12.7 & 1.0 &  0.8 &  57.0 & 93.0 & 16 &	943 &	831 & 77.3 & 79.5\\ 
		one-to-one & 16.4 & 1.0 &  0.6 &  46.1 & 93.3 &  5 &	1746 &	491 &  58.0 & 83.1\\ 
		\hline
		\hline
	\end{tabular}
\end{table*}

\begin{table*}[!t]
	\renewcommand{\arraystretch}{1.3}
	\caption{Consuming Time(Second) Comparison Between Our Tracking Method and Others Methods for the $Larva$ Dataset}
	\label{table_time_comparison}
	\centering
	\begin{tabular}{c|c c c c}
		\hline
		\hline
		\textbf{Method} & \textbf{larva\_s1} & \textbf{larva\_s2}&\textbf{larva\_s3} & \textbf{Sum} \\
		\hline
		Ours & 2.5 & 4.4 & 2.1 & 9.0 \\
		Cox's & 109.8 & 310.5 & 45.2 & 465.5 \\
		\hline
		\hline
	\end{tabular}
\end{table*}

\begin{figure*}[!t]
	\centering
	\includegraphics[width=7.0in]{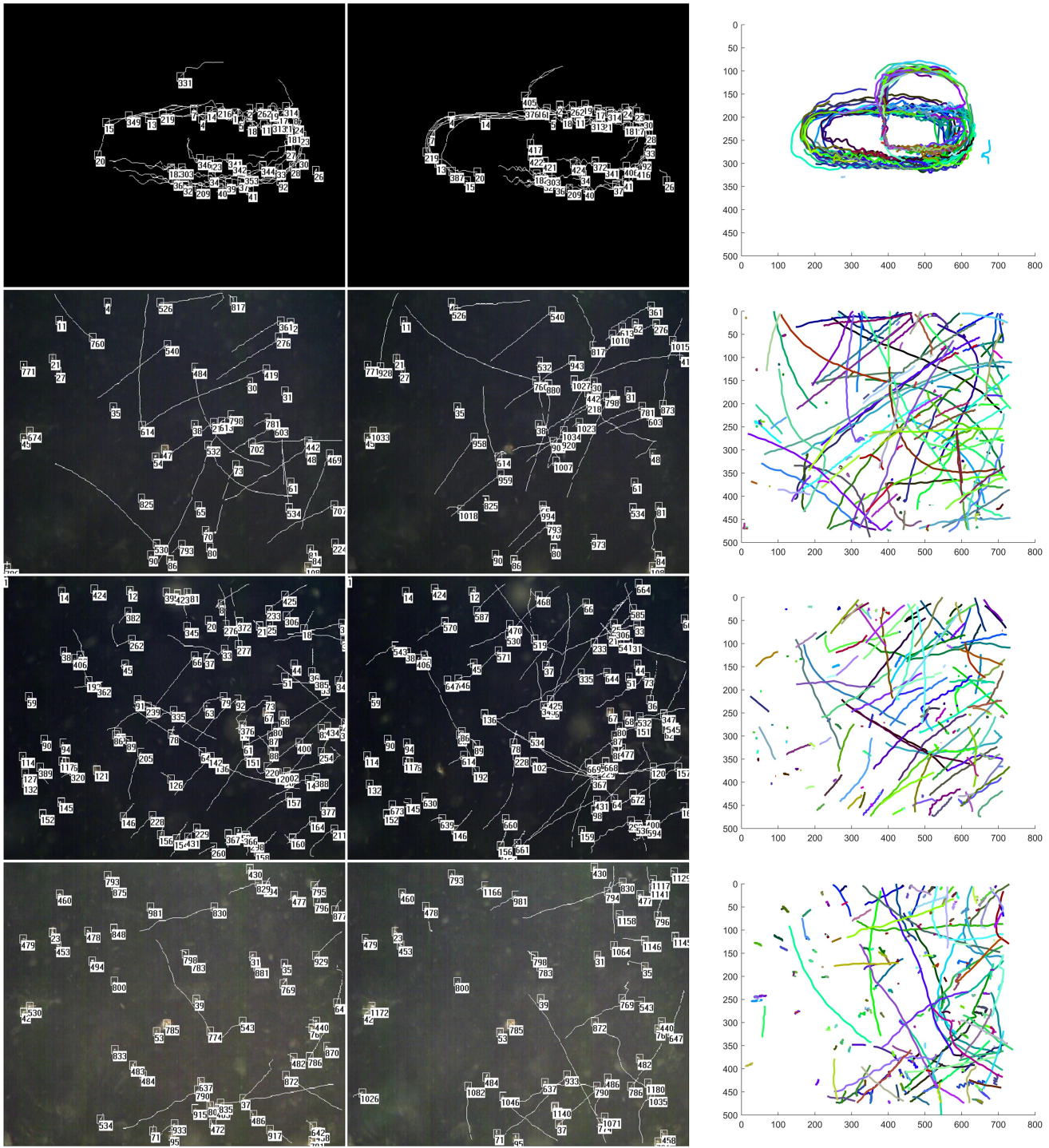}
	\caption{Results on the datasets $Larva$ and $Verti\_Hat$. Each line of the pictures shows the tracking results in a sequence clip, including two frame images(last frame and twentieth from the end) and a global projection graphic displaying all trajectories. Object ID and trajectories are marked on the image. Only last 30 frames of trajectories are drawn to keep the picture more readable. Square boxes indicate the position of small object in current frame. From top to bottom, four lines of pictures indicate the $hat$,$Larva\_s1$,$Larva\_s2$ and $Larva\_s3$ respectively.}
	\label{fig_track_result}
\end{figure*}

\paragraph{Experiment of one-to-many assumption}
We also provide a comparison test to prove the effect of our one-to-many assumption. By using different constraints for track selection, the results of flexible association (one-to-many) and hard association (one-to-one) are listed as table \ref{table_tracking_larva_s1_one_to_one}. One-to-one association refuses some potential hypotheses, because that 'to be or not to be' decision means compromised losing in some cases.

As Tables \ref{table_tracking_larva_s1_one_to_one} shows, our one-to-many constraint boosts the performance notably. Critical indexes like MOTA,OSPA-T are improved by a quite large range. OSPA-T decreases by 22\%. Meanwhile MOTA increases by more than 20\%. However, some metrics appear modest degeneration such as IDSW. Even though, the one-to-many constraint is significative, in consideration of obvious improvement on critical indexes.

During MMHT stage, our efforts to keep the number of hypotheses in a tractable scale have removed most short and weak hypotheses. Leaving the strong and long hypotheses, at this point, too rigorous constraint could reject imperfect ones with small flaws. In the extended time duration and complex scenes, some strong hypotheses inevitably encounter long occlusion crossing like we illustrated before, where measurement competition occurs. One-to-one association will reject a strong hypothesis for such flaws while the major part of the hypothesis is correct. Then, since we already removed those weak short hypotheses before, no succedaneum composed of splitting hypotheses can be used as an inferior choice. So, our one-to-many constraint complements this procedure gap and preserves the practicable strong hypothesis.

One-to-many constraint is compatible with our tracking framework. In addition, one-to-many constraint probably could be a new direction that breaks through the traditional idea of hard association from a certain aspect. We only provide modest or slight relaxation for one-to-many constraint, only limited number of sharing measurements is permitted. Massive number of sharings will entangle the tracking problem and degenerate the tracker performance. 

\paragraph{Experiment of speed comparison}
To save consuming time is another important target for tracking, especially for motion analysis of dense object. However, this is often regarded as an extra factor for the extreme difficulty and the tradeoff between performance and speed. For our tracking method, we achieve decent performance. Meanwhile consuming time is restricted under a pretty low level. As the Table \ref{table_time_comparison} show, our tracking method is more than ten times faster than Cox's MHT. The platform we used is a PC with E5(3.3GHz) CPU. Only two methods are listed in the table, because they use the same language(C++) while others use Matlab. Cox's MHT algorithm possesses both well tracking capability and efficient implementation, according to the recent comparison in Chanho Kim \textit{et.al} \cite{MHTICCV}. From some respects, Cox's implementation remains a meaningful reference.
%
The fast implementation is based on our aiming to screen out weak hypothesis at first place and preserve strong ones later, cooperating with batch optimization instead of evolution frame by frame.

\section{Conclusion}

In the paper, we propose and implement the methods for an entire tracking processing including detection and tracking for dim small object, from raw image sequence to identified object trajectories. For the detection, we present our multi-appearance segmentation. It employs multi-layer thresholds to produce multi-appearance slides, and exploits the tree structure to describe the relations between objects from different layers. Instead of global segmentation, we try to achieve one threshold for one detection, to maximize the adaptability. Then, a deep-first-branch-adjustment algorithm is designed to solve the optimization of threshold for every individual object. According to the final tracking result, this detection method markedly improved the performance of our tracking method, providing valued detection input. 
 
We build the tracking management structure based the classical tree from TOMHT. Then we implement it with some heuristic techniques, such as loop detect based hypothesis merge, modified motion score utilizing the auto-regression information, etc. For the global hypothesis selection, we propose a extended 0-1 program based on the idea of one-to-many association, integrating compatibility information and object likelihood in the meantime. This new idea permits fewer number of sharings that one measurement can be assigned to multiple objects. This means would improve the "cut-off" phenomenon and preserve the identity of objects. Thanks to our efforts to reduce complexity and number of hypotheses, our tracking method is implemented in an efficient way. It's proved to be impressively fast in experiments.


%

\appendices

\section*{Acknowledgment}

We would like to thank Prof. Han Sun from Department of Electrical and Computer Engineering, Nanjing University of Aeronautics and Astronautics, who provided part of video in experiment data.

\ifCLASSOPTIONcaptionsoff
  \newpage
\fi



%
\bibliographystyle{IEEEtran}
\bibliography{sot}

\end{document}